\documentclass[twocolumn]{article}
\pdfoutput=1

\usepackage[margin=0.8in]{geometry}
\usepackage{times}
\usepackage{epsfig}
\usepackage{graphicx}
\usepackage{amsmath}
\usepackage{amssymb}
\usepackage{color}
\usepackage{makecell}
\usepackage{caption}
\usepackage{subcaption}
\usepackage{titling}

\usepackage[ruled]{algorithm2e}
\usepackage{algorithmic}
\SetAlCapHSkip{0pt}
\usepackage[pagebackref=true,breaklinks=true,colorlinks,bookmarks=false]{hyperref}

\renewcommand{\tt}[1]{\texttt{#1}}

\DeclareMathOperator*{\argmin}{arg\,min}

\date{}

\begin{document}

\title{ \textbf{GPU-Based Homotopy Continuation for Minimal Problems in Computer Vision}}

\author{Chiang-Heng Chien\\
School of Engineering\\
Brown University\\
{\tt\small chiang-heng\_chien@brown.edu}
\and
Hongyi Fan\\
School of Engineering\\
Brown University\\
{\tt\small hongyi\_fan@brown.edu}
\and
Ahmad Abdelfattah\\
Innovative Computing Laboratory\\
University of Tennessee\\
{\tt\small ahmad@icl.utk.edu}
\and
Elias Tsigaridas\\
INRIA\\
{\tt\small elias.tsigaridas@inria.fr }
\and
Stanimire Tomov\\
Innovative Computing Laboratory\\
University of Tennessee\\
{\tt\small tomov@icl.utk.edu}
\and
Benjamin Kimia\\
School of Engineering\\
Brown University\\
{\tt\small benjamin\_kimia@brown.edu}
}

\maketitle

%\thispagestyle{empty}

%%%%%%%%% ABSTRACT
\begin{abstract}
Systems of polynomial equations arise frequently in computer vision, especially in multiview geometry problems.  Traditional methods for solving these systems typically aim to eliminate variables to reach a univariate polynomial, {\em e.g.,} a tenth-order polynomial for 5-point pose estimation, using clever manipulations, or more generally using Grobner basis, resultants, and elimination templates, leading to successful algorithms for multiview geometry and other problems. However, these methods do not work when the problem is complex and when they do, they face efficiency and stability issues. Homotopy Continuation (HC) can solve more complex problems without the stability issues, and with guarantees of a global solution, but they are known to be slow. In this paper we show that HC can be parallelized on a GPU, showing significant speedups up to 26 times on polynomial benchmarks. We also show that GPU-HC can be generically applied to a range of computer vision problems, including 4-view triangulation and trifocal pose estimation with unknown focal length, which cannot be solved with elimination template but they can be efficiently solved with HC. GPU-HC opens the door to easy formulation and solution of a range of computer vision problems.
\end{abstract}

\section{Introduction}
Systems of polynomial equations arise frequently in computer vision, especially in multiview geometry problems, because perspective projection is an algebraic model. Examples abound including 
absolute pose estimation~\cite{Haner2015cvpr, wu2015cvpr, albl2015cvpr}, 
relative pose estimation~\cite{nister2004efficient, henrik2005relativepose, kuang2014cvpr}, 
pose estimation with unknown focal length~\cite{bujnak2008cvpr}, homography estimation~\cite{kukelova2015cvpr, brown2007minimal}, PnP~\cite{zheng2013revisiting, zheng2014cvpr}, 3-View triangulation~\cite{byrod2007fast}, 
pose estimation with unknown principal point~\cite{larsson2018camera}, rolling shutter camera absolute pose estimation~\cite{albl2019rolling}, as well as many others. The challenge has been how to solve these polynomial systems efficiently and in a stable way. 

The classic 5-point algorithm for relative pose estimation~\cite{philip1996non,nister2004efficient} is a case in point. Its formulation begins with 15 equations in 15 unknowns, namely, 10 depths and 5 pose parameters. The traditional approach is to eliminate depths and end up with the epipolar equation which with 5 points results in a 10th-degree univariate polynomial from which pose is determined. A more formal approach to eliminating variables is the Gr\"obner basis ~\cite{cox2013ideals,cox_using_2005} or resultants~\cite{cox2013ideals,cox_using_2005}. Elimination Templates were developed as an automatic solver generator~\cite{larsson2017efficient} where the Gr\"obner-based elimination strategy obtained from one input is ``remembered" for future inputs. These methods are reviewed in Section~\ref{sec:methods_poly}. \\
\indent The challenge with the above methods is that they are limited to problems with small number of solutions. They are slow for larger problems whose elimination template can be computed. For even larger problems the computation of elimination template exceeds practical resources, rendering the problem unsolvable. In addition, stability issues might arise in the process of converting a system of polynomials to a single univariate polynomial, e.g., \cite{mourrain2012border,mourrain2007pythagore}. \\
\indent Homotopy Continuation methods, in contrast, can solve very complex polynomial systems. The basic idea is to find all the solutions of a start  system and then to continuously evolve them to the solutions of the target system. They can ensure, with probability 1, to find all solutions \cite{sommese2005numerical,verschelde1999algorithm}, provided a ``good" starting system. They also avoid the stability issues of symbolic methods as they do not manipulate the input polynomials. Their complexity depends on the number of solutions (tracks) they follow. In this lies the idea to use a GPU to speed up the computation. \\
\indent GPUs have been used in computer graphics and computer vision to accelerate massively parallel operations. The key is whether HC can be parallelized to take advantage of many processor in a GPU while avoiding data transfer delays. The HC process consists of prediction and correction steps in the continuation from the start system to the target system. This is done by computing the Jacobian to predict where to go next, and subsequently Newton’s method to correct the solution. We show that by parallelizing the computations in the prediction and correction steps, a track can be implemented on a \emph{warp}. This is made possible in part by instituting kernel fusion in the MAGMA library for solving batch linear systems. In addition, an indexing system homogenizes the expressions of Jacobian and the two vectors involved so their evaluations can be parallelized. The resulting GPU-HC can be generically applied to systems of up to 32 equations by 32 unknown and speedups of up to 26 times on polynomial benchmarks. \\
\indent Computer vision problems involving polynomial systems fit these requirements. We have applied GPU-HC to a variety of problems, and found that for moderately complex systems and beyond GPU-HC offers significant savings (with implied stability). We have also explored solutions to two problems, namely, 4-view triangulation and trifocal pose estimation with unknown focal length which have not been explored in the literature. These are introduced as example cases where elimination template fails to produce solutions but GPU-HC solves efficiently. The basic thesis of this paper is that GPU-HC can be applied to all computer vision problems that can be formulated as polynomial systems and produce efficient and stable solutions.

\section{Methods for Solving Polynomial Systems}
\label{sec:methods_poly}

We partition the algorithms for solving systems of polynomial equations in roughly three categories: {\em (i)} \textbf{Symbolic methods} that rely on algebraic
elimination tools, such as Gr\"obner basis, resultant, {\em etc.}; {\em (ii)}
\textbf{Numerical solvers} that are iterative and are generally a variant of Newton's method, such as homotopy continuation, and/or rely on eigenvalue computations; and, {\em (iii)}
\textbf{Hybrid methods} that combine the benefits of the symbolic and numerical solvers such as elimination templates or subdivision solvers. 
% Below each type of approach is covered in more detail.

\noindent \textbf{Symbolic solvers} ``transform", using algebraic
elimination, the multivariate polynomial system to a univariate
polynomial. The roots of this polynomial are computed using
dedicated algorithms, like Sturm or Descartes, and are used to recover the system solutions, \emph{e.g.},
\cite{cox2013ideals,cox_using_2005,rouillier_solving_1999,elkadi_introduction_2007}.
These algorithms mainly rely on exact computations with rational numbers and partially on computations in finite fields. They perform elimination using
well-known tools from computational algebraic geometry, such as Gr\"obner basis
and resultants. Gr\"obner basis manipulate the polynomials ``incrementaly"
(like Gaussian elimination) to deduce the univariate
polynomial, while resultants use all the polynomials right from the beginning (similar to Cramer's rule).

Symbolic methods are used widely in solving minimal problems in computer
vision~\cite{kneip2012finding, kneip2013direct, fabbri2020camera,
  stewenius2005grobner,henrikstewenius2005solutions}. They always
output the exact results with certifications. They deal successfully and rather
efficiently with degeneracies such as multiple roots. The efficient
implementation of symbolic algorithms is far from a straightforward task;
various sub-algorithms must be fine-tuned, implemented, and extended
experimentation is needed. However, despite the tremendous recent progress in
this direction, systems of more than 5-6 variables of moderate degrees cannot be
handled, except if sparsity and the structure 
% of the underlying system
is specifically exploited. Even more, we are still very far from having symbolic
solvers that solve moderate systems in milliseconds. \\
\indent Another major issue with symbolic solvers, especially Gr\"obner basis, is that they are numerically unstable~\cite{kreuzer2000computational,mourrain2007pythagore}.
This is mainly due to their requirement for a term-ordering 
that causes instability when the coefficients of the input polynomials
are floating point numbers or known up to some precision. 
Nevertheless, there are efforts to overcome this obstacle
using a variant called border basis, {\em e.g.}, \cite{mourrain2012border}.
The same phenomena appear in the resultant computations
\cite{noferini2016numerical}, where there also recent efforts 
for improvements \cite{bender2021yet}.

\noindent \textbf{Numerical solvers} are almost exclusively iterative algorithms 
that exploit a variant of Newton operator
and they perform their computations in floating point arithmetic, e.g.,~\cite{Bertini-book,sommese2005numerical,verschelde1999algorithm}.
There are also approaches based on numerical linear algebra techniques, mainly on eigenvalue computations e.g., \cite{bender2021yet,buse2005resultant}.
The most prominent representatives are the Homotopy Continuation (HC) algorithms
~\cite{alexander1978homotopy, bates2008adaptive, chen2014hom4ps, holt1990experience,verschelde1999algorithm,hauenstein2018adaptive}.
They rely on the simple and elegant idea
to initially solve a simpler polynomial system (start system)
and then deform its roots to 
the roots of the system we want to solve (target system). 
Some care is required on choosing an easy-to-solve start system
that has at least as many solutions as the target system. 
They can handle very big problems,
especially in the absence of degeneracies, say multiple roots.
These solvers are highly efficient in practice and able to handle systems that are out of the reach of symbolic solvers. Nevertheless, they are still comparatively slow, a serious bottleneck to their wide adoption. Their potential for parallelization is a key focus of this paper. 
HC is used widely in computer vision,
%for solving polynomial systems, 
especially for minimal problems in multiview geometry~\cite{kriegman1992geometric, pollefeys1997vnl, maybank1992theory, duff2020pl,fabbritrifocal, duff2019plmp}.

Numerical problems might also occur in HC 
algorithms, especially if the Jacobian of the system is ill-conditioned and in the many cases we need to 
use double-precision floating point arithmetic, 
{\em e.g.}, \cite{Bertini-book}. However, HC
is an inherit numerical method and does not require an exact input.
%the assumption
%that we know exactly the coefficients of the input polynomials.

Also sometimes it is not easy, if possible at all, to find good, let alone optimal, start systems, the cardinality of the output is not always correct,
and extra verification steps are needed.
They are in general easier to implement than symbolic methods, even though in all the cases efficient scientific software requires tremendous amount of time, energy, and effort to be efficient and solve real life problems. 

\noindent \textbf{Hybrid solvers} aim to combine the symbolic
and numerical approaches
e.g.~\cite{elkadi2005symbolic,mourrain2007pythagore,mantzaflaris2011continued}, and they have various algorithmic variants. A well-known
 method in the computer vision community is the "elimination template", or automatic solver generation~\cite{kukelova2008automatic,
  kukelova2017clever, larsson2018beyond, li2020gaps, larsson2017efficient}. The
main idea is to bookkeep the steps that an elimination (usually Gr\"obner basis)
algorithm performs for one input and apply these steps to any other input. They
generate a ``template" of elimination at an offline stage with the random
coefficients of a ``dummy" system on a finite field. 
%Then, we can do elimination
%for any other input system by performing (numerical) linear algebra, typically
%Gaussian-Jordan elimination and/or linear system solving. 
We obtain
the solutions by eigenvalue computations or dedicated algorithms. The
method is particularly fast for solving systems with low degree and low number
of variables~\cite{pritts2018radially, chen2018polarimetric, zhao2019minimal,
  larsson2019revisiting, albl2019rolling, ding2020efficient}.

Nevertheless, even though they have turned out to be quite successful in some problems, they cannot always guarantee their result, 
they might also need to handle very large matrices \cite{larsson2017efficient} which are computationally intractable, and, last but not least, it is far from trivial to analyze their stability.
The hybrid approaches based on elimination template method 
try to overcome the instability of the symbolic methods by performing 
several pre-computations. However, at the end they also must compute 
with a matrix, similar to Gr\"obner basis and resultants,
which has a dimension at least the number of complex solutions.
The condition number of such matrix is not well, if at all, studied, and it is not clear if they can handle problems with
$\geq 300$ roots.

\section{Homotopy Continuation}
\label{sec:HC}
The idea of Homotopy Continuation (HC)~\cite{morgan2009solving,sommese2005numerical} is to evolve the solutions of one polynomial system $G$, the ``start system", to discover the solutions of another system $F$. Let $X = (x_1, x_2, ... ,x_M)$ represent $M$ unknowns. Let $F(X)$ be a system of $N$ polynomial equations $F = (f_1,f_2, ... , f_N)$; this is the ``target system" we want to solve. Let $G(x)$, $G = (g_1,g_2, ... , g_N)$ be the ``start system" whose solutions are all known. The idea of HC is to construct a series of intermediate polynomial systems $H(X,t)$, $H = (H_1,H_2, ... , H_N)$; where $H(X,0) = G(X)$ and $H(X,1) = F(X)$, {\em e.g.}, via linear interpolation:
\begin{align}
    H(x,t) = (1 - t)G(x) +  tF(x), \qquad t \in [0, 1].
\label{Eq:HC}
\end{align}

\begin{figure}
    \centering
    \includegraphics[width = 0.9\linewidth]{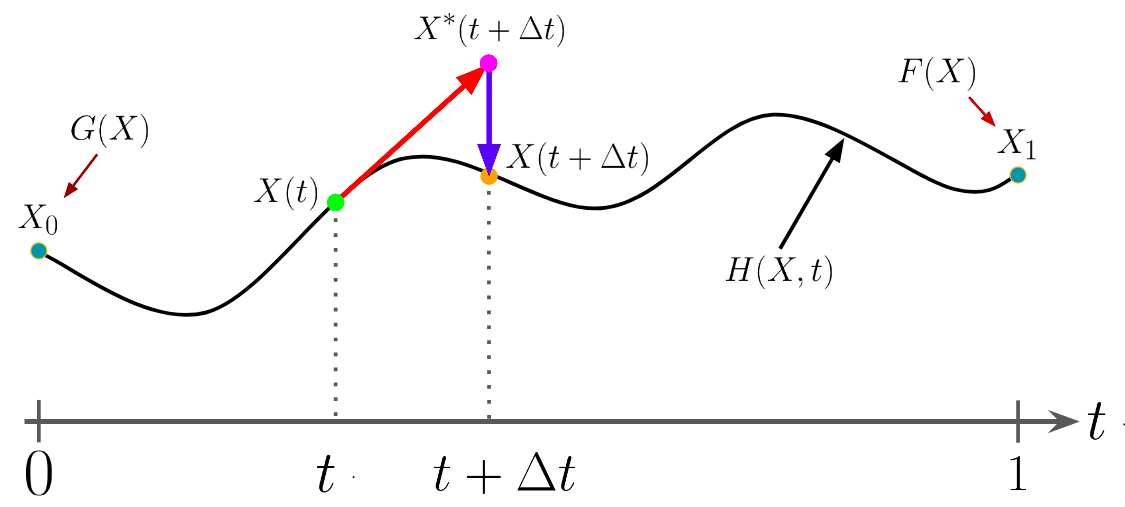}
    \caption{A track (curve) of a Homotopy Continuation algorithm showing $H(X,t)$ in black, along with one prediction (red) and one correction (blue).}
    \label{fig:HCIdea}
\end{figure}

The basic idea is to find the solution of $H(X,t+\Delta t)$ from the solution of $H(X,t)$. Figure~\ref{fig:HCIdea} illustrates the idea for one solution and one unknown. The black curve is the locus of the solution $X(t)$ of $H(X,t)$, the homotopy curve, where $X_0$ is the known solution of $G(X)$ and $X_1$ is the desired solution of $F(X)$. We track solution $X_1$ from $X_0$ in a number of small steps, each consisting of a prediction and a correction step. Prediction uses a first-order Taylor expansion to estitmate $X$ at $t+\Delta t$ in the form of  
\begin{align}
    X^*(t + \Delta t) = X(t) + \tfrac{d X}{d t} \Delta t,
\end{align}
where $X^*$ is the first order estimation of $X(t + \Delta t)$. 
We obtain $\frac{dX}{dt}$ by differentiating $H(X(t),t)$, {\em i.e.},
\begin{align}
\label{Eq:linsysRK}
     \frac{\partial H}{\partial X} \frac{d X}{d t} + \frac{\partial H}{\partial t} = 0 \longrightarrow \frac{d X}{d t} = -(\frac{\partial H}{\partial X})^{-1}\frac{\partial H}{\partial t},
\end{align}
where $J =  \frac{\partial H}{\partial x}$ is the $M \times N$ Jacobian of $H$ wrt $X$, giving 
\begin{align}
\label{Eq:prediction}
    X^* (t + \Delta t) = X(t) - (\tfrac{\partial H}{\partial X})^{-1}\tfrac{\partial H}{\partial t} \Delta t.
\end{align}
This step, the first-order estimation of $X^*$ from $X(t)$, is known as the \textbf{prediction} step (Figure~\ref{fig:HCIdea}). However, we can improve the prediction using a higher-order method like a 4-th order Runge-Kutta; alas, we require a \textbf{correction}. Using Newton we update $X^*(t+ \Delta t)$ to $\hat{X}(t + \Delta t)$, {\em i.e.}, 
\begin{align}
    H(X^*,t+ \Delta t) + \tfrac{\partial H}{\partial X}(X^*,t+ \Delta t)(\hat{X} - X^*) = 0,
     \label{Eq:homotopy}
\end{align}
giving the estimate $\hat{X}$ in the form of 
\begin{align}
    \hat{X} =  X^* - (\tfrac{\partial H}{\partial X})^{-1}(X^*,t+ \Delta t) H(X^*,t+ \Delta t).
    \label{Eq:newton}
\end{align}
This is the correction step. The pairs of prediction and correction steps numerically evolve $X_0$ as the solution of $G(X)$ to $X_1$ as the solution of $F(X)$.

Provided that we have a good started system the HC algorithms find all the solutions (up to some approximation) with probability one. Even more, there are methods, alas much slower, that can guarantee that we follow accurately the tracks \cite{beltran2013robust} and/or certify the solutions \cite{hauenstein2012algorithm}.

% Homotopy Continuation (HC) is a (family of) numerical algorithm(s) for solving 
% a system of equations $F(x)$ , where $F = (f_1,f_2, ... , f_N)$ are polynomials in $M$ unknowns, $x = (x_1, x_2, ... ,x_M)$m with complex coefficients. 

\section{Illustrative Problems}
\label{sec:illustractive_problems}
% Several classic computer vision problem illustrate how HC can be used to solve them efficiently. THis is then used to solve two problem which cannot be tackled with traditional approachs. 

\noindent \textbf{Preliminaries:} Let $\Gamma$ denote a 3D point which projects to an image point $\gamma = (\xi, \eta, 1)^T$ with depth $\rho$ so that $\Gamma = \rho \gamma$. The expression of $\Gamma$ in a camera related by pose ($R$,$T$) to another camera where $R$ is the rotation matrix and $T$ is translation, is $\bar{\Gamma} = R \Gamma + T$. Due to metric ambiguity the unit direction $\hat{T}$ along $T$ in sought, where $T = \lambda \hat{T}$. 

\noindent \textbf{Relative Pose Estimation with Calibrated Cameras} is a classic problem most frequently solved by Nister's 5-point algorithm~\cite{nister2003efficient,nister2004efficient,philip1996non}. Consider five corresponding points ($\gamma_i$, $\bar{\gamma}_i$) where $\gamma_i$ in one image is in correspondence with $\bar{\gamma}_i$ in the second image. Since $\Gamma_i = \rho_i \gamma_i$, $\bar{\Gamma}_i = \bar{\rho}_i \bar{\gamma}_i$, and $\bar{\Gamma}_i = R \Gamma_i + T$. The relationship between $\gamma_i$ and $\bar{\gamma}_i$ is captured as
\begin{align}
    \bar{\rho}_i \bar{\gamma}_i = R \rho_i \gamma_i + \hat{T}, \qquad  i=1,2,\cdots,N,
\label{Eq:FullRigidMotion}
\end{align}
where the depths ($\rho_i$,$\bar{\rho}_i$) represent 10 unknowns and ($R$,$\hat{T}$) represent 5 unknowns. The above set of five vector equations give 15 constraints in 15 unknowns. Representing $R$ with quaternions which involves 4 unknowns with one equation yields 16 {\em polynomial equations} in 16 unknowns. Observe that there has been no attempts in the literature to solve these equations, which HC can solve, referred to as relative pose estimation $+$ depth reconstruction in Table~\ref{tab:VisionProblems}. Rather, the traditional approach is to reduce the number of unknowns by eliminating the ten depth variables by taking cross product of Equation~\ref{Eq:FullRigidMotion} with $\hat{T}$ and then dot product with $\bar{\gamma}_i$ giving the classical epipolar relationship 
\begin{align}
    \bar{\gamma}_i^T E \gamma_i = 0,   \qquad  i=1,2,\cdots,5.
\label{Eq:epiConstraint}
\end{align}
where $E = [\hat{T}]_{\times} R$. While this is now 5 equations in 5 unknowns ($R$,$\hat{T}$), these now involve trigonometric equation unless R is represented with a quaternion giving 6 polynomial equations in 6 unknowns. Again, this can also be solved by HC. Nevertheless the classic approach is to treat $E$ as nine unknowns and use a Theorem~\cite{nister2003efficient} that $E = [\hat{T}]_{\times} R$ if and only if 
\begin{align}
    2 E E^T E - trace(EE^T)E = 0.
\label{Eq:traceConstraint}
\end{align}
These are 9 cubic polynomial equations but only four are independent which can be used in conjunction with Eq.~\ref{Eq:epiConstraint} to solve for $E$. Namely, $E$ is written in vector form as $\tilde{E}$,
{\footnotesize
\begin{align}
\left\{ \begin{matrix}
\tilde{E}^T= [E_{11}, E_{12}, E_{13}, E_{21}, E_{22}, E_{23}, E_{31}, E_{32}, E_{33}]\\
    w^T_i \tilde{E} = 0, \qquad i = 1,2,\cdots,5, \\
    w^T = [\xi_i\bar{\xi}_i, \eta_i\bar{\xi}_i, \bar{\xi}_i, \xi_i \bar{\eta}_i, \eta_i \bar{\eta}_i, \bar{\eta}_i, \xi_i, \eta_i, 1].	
    \end{matrix}
\label{Eq:linearSystem}
\right. 
\end{align}
}
$\tilde{E}$ is then an arbitrary linear sum of the four matrices representing the right nullspace, $\tilde{E} = \alpha_1 E_1 + \alpha_2 E_2 + \alpha_3 E_3 + E_4$, where the last constant $\alpha_4$ is set to one due to the scale invariance of $E$. The only remaining constraint is the set of nine cubic Equations~\ref{Eq:traceConstraint}, where the unknowns ($\alpha_1$, $\alpha_2$, $\alpha_3$) involve 20 monomials up to order 3 of ($\alpha_1$, $\alpha_2$, $\alpha_3$), so that they can be expressed as a $9 \times 20$ matrix multiplied by a vector or 20 monomials. The idea is to eliminate all monomials except those involving one variable, say $\alpha_3$. This can be done by Gauss-Jordan elimination with partial pivoting to make an upper triangular matrix, and after additional manipulations, which are effectively hand-derived Gr{\"o}bner basis, leads to a single tenth-order polynomial in one variable $\alpha_3$ which gives 10 roots. The real roots of $\alpha_3$ then can solve for $\alpha_1, \alpha_2$ and $E$ from which $R$ and $T$ can be recovered.

Li and Hartley~\cite{li2006five} solve Equation~\ref{Eq:traceConstraint} with $\tilde{E}$ as described by Equation~\ref{Eq:linearSystem} using the hidden variable technique, a resultant technique for algebraic elimination~\cite{cox2013ideals}. They include $det(E)=0$ as a tenth equation and solve equating the determinant of the $10 \times 10$ matrix to zero as a function of $\alpha_3$, a tenth-order polynomial which can then be solved. The claimed advantage of this technique over Nister’s is its simplicity and ease of implementation. 

Observe that both approaches devise ingenius algorithms to turn the basic system of polynomial Eqaution~\ref{Eq:FullRigidMotion} into a single 10th degree uni-variate polynomial. In contrast, Homotopy Continuation can be used immediately to solve as $16 \times 16$ polynomial system or the reduced $6\times6$ of system of Equation~\ref{Eq:traceConstraint} avoiding the need for devising such algorithms. Finally, HC can be used to solve ($\alpha_1$, $\alpha_2$, $\alpha_3$) using a $3 \times 3$ system of cubic polynomials. Note that we are not advocating to solve the relative pose using HC (the system is too small to benefit from it). Rather, we are noting that it \emph{can} be solved by HC as an illustration.

\noindent \textbf{Perspective-n-Point problem (PnP)}  estimates the pose of a calibrated camera ($R$,$T$) using $n$ correspondences between 3D world coordinate points $\Gamma_i$  and their 2D projections in the image $\gamma_i$ (known as space resection in photogrammetry). The P3P problem where 3D points ($\Gamma_i$, $\Gamma_2$, $\Gamma_3$) correspond to 2D image points ($\gamma_1$, $\gamma_2$, $\gamma_3$), respectively, has a long history~\cite{grunert1841pothenotische,finsterwalder1903ruckwartseinschneiden, haralick1991analysis,quan1999linear} and it has 4 solutions requiring a 4th correspondence to disambiguate. 

The basic formulation can be posed using $\Gamma_i = \rho_i \gamma_i$ as
%are in correspondence with three 2D image points $\gamma_1$, $\gamma_2$, and $\gamma_3$, respectively, to determine the pose of the camera with respect to the world in terms of (R,T). The traditional approach is to first obtain the three depth unknowns $\rho_i$ defined from $\Gamma_i =\rho_i \gamma_i$. Observe that the cosine relationship in a triangle connecting the camera center to 3D points $\Gamma_i$ and $\Gamma_j$ relates the 
% $|\Gamma_i - \Gamma_j|^2 = |\Gamma_i|^2 + = |\Gamma_j|^2 -2 |\Gamma_i| |\Gamma_j| \cos (\angle \Gamma_i,O,\Gamma_j)$,
% Where $O$ is the camera center. Since the angle also be determined by a cosine relation in the triangle connecting $\gamma_i$, $O$ and $\gamma_j$ using
% $|\gamma_i - \gamma_j|^2 = |\gamma_i|^2 + = |\gamma_j|^2 -2 |\gamma_i| |\gamma_j| \cos (\angle \Gamma_i,O,\Gamma_j)$,
% The cosine can be found and the equations are only in terms of $\rho_i$, $\rho_j$. The following is a cleaner exposition:
% The following is a geometric approach to the standard P3P problem. Consider three point correspondence between image points $\{\gamma_1, \gamma_2, \gamma_3\}$, with 3D points $\{\Gamma_1, \Gamma_2, \Gamma_3\}$, respectively, with unknown depth $\{\rho_1, \rho_2, \rho_3\}$. Then, if the camera pose is denoted by the rotation matrix $R$ and translation vector $T$, we have
\begin{align}
    \left \{
    \begin{matrix}
                \Gamma_1 = \rho_1 R \gamma_1 + T \\
                \Gamma_2 = \rho_2 R \gamma_2 + T \\
                \Gamma_3 = \rho_3 R \gamma_3 + T
    \end{matrix}
    \right.
\label{Eq:P3POriginal}
\end{align}

\noindent a set of nine equations in the nine unknowns. At this point where the formulation is completed, HC can be used to solve for ($R$,$T$), as well as depthes! Using a quaternion representation of $R$ which involves 4 unknowns and one equation, this becomes a set of $10 \times 10$ polynomials with 10 unkowns. The traditional approach eliminates $R$ and $T$ to solve depth from

%\begin{align}
%    \left \{
%    \begin{matrix}
%                \Gamma_2 - \Gamma_1 = R (\rho_2 %\gamma_2 - \rho_1 \gamma_1)\\
%                \Gamma_3 - \Gamma_1 = R (\rho_3 %\gamma_3 - \rho_1 \gamma_1)
%    \end{matrix}
%    \right.
%\end{align}
%and eliminating the rotation matrix through a pairwise dot product:
{\footnotesize
\begin{align}
    \left \{
    \begin{matrix}
     (\Gamma_2 - \Gamma_1)^T (\Gamma_2 - \Gamma_1) = (\rho_2 \gamma_2 - \rho_1 \gamma_1)^T(\rho_2 \gamma_2 - \rho_1 \gamma_1)\\
     (\Gamma_3 - \Gamma_1)^T (\Gamma_3 - \Gamma_1) = (\rho_3 \gamma_3 - \rho_1 \gamma_1)^T(\rho_3 \gamma_3 - \rho_1 \gamma_1) \\
     (\Gamma_2 - \Gamma_1)^T (\Gamma_3 - \Gamma_1) = (\rho_2 \gamma_2 - \rho_1 \gamma_1)^T(\rho_3 \gamma_3 - \rho_1 \gamma_1)
     \end{matrix}
    \right.,
\label{Eq:P3PafterElim}
\end{align}
}

\noindent a set of three quadratic in three unknowns ($\rho_1$,$\rho_2$,$\rho_3$). Again, this reduced form can be easily solved by HC, but the traditional approach is to apply 
% {\footnotesize
% \begin{align}
%     \left \{
%     \begin{matrix}
%         \gamma_1^T \gamma_1 \rho_1^2 - 2 \gamma_1^T \gamma_2 \rho_1 \rho_2 + \gamma_2^T \gamma_2 \rho_2^2 = (\Gamma_2 - \Gamma_1)^T (\Gamma_2 - \Gamma_1) \\
%         \gamma_1^T \gamma_1 \rho_1^2 - 2 \gamma_1^T \gamma_3 \rho_1 \rho_3 + \gamma_3^T \gamma_3 \rho_3^2 = (\Gamma_3 - \Gamma_1)^T (\Gamma_3 - \Gamma_1) \\
%         \gamma_1^T \gamma_1 \rho_1^2 - \gamma_1^T \gamma_2 \rho_1 \rho_2 - \gamma_1^T \gamma_3 \rho_1 \rho_3 + \gamma_2^T \gamma_3 \rho_2 \rho_3 = \\ (\Gamma_2 - \Gamma_1)^T (\Gamma_3 - \Gamma_1)
%     \end{matrix}
%     \right.
% \end{align}
% }
% These equations can be solved by applying the 
Silvester’s resultant to get an 8th degree polynomial, containing even terms so that it is effectively a quartic~\cite{quan1999linear}. 
%Observe that for larger system eliminating many equations in to a single univariable polynomial is unstable~\cite{kreuzer2000computational,mourrain2007pythagore}. 

The general PnP problem relies on $n$ correspondences between 3D points $\Gamma_i$ and 2D image points $\gamma_i$, $i=1,2,\cdots, n$. A direct minimization of the algebraic reconstruction error~\cite{zheng2013revisiting} uses a non-unit quaternion representing of $R$ and explicitly optimize for R. This gives four polynomials of degree three in four variables, which are solved by Gr{\"o}bner bases, from which an elimination template is constructed using the automatic generator in~\cite{kukelova2008automatic}. This gives at most 81 solutions with an 575x656 elimination template and 81x81 action matrix. Alternatively, these equations can be solved using HC without any further processing with about a factor of 5 times speedup on a GPU, Table~\ref{tab:VisionProblems}. In this larger case, HC features both simplicity and effciency. 

\noindent \textbf{N-view Triangulation} aims to find the 3D world point $\Gamma$ that is most consistent with a set of projection, $\gamma_1,\cdots, \gamma_N$ from $N$ views, given relative pose of all cameras in the form of the pairwise essential matrix $E_{ij}$ between views $i$ and $j$. Due to noise, the projection rays from corresponding points do not necessarily meet in space. For two cameras, the mid-point between the closest points on the projection rays is used~\cite{beardsley1994navigation}. But this can have a large error, especially with large calibration error. Rather then minimizing the latent 3D error, the reprojection error can be minimized~\cite{hartley1995triangulation,hartley1997triangulation,kanatani2008triangulation}. Let $\gamma_i = \hat{\gamma}_i + \Delta \gamma_i$ where $\hat{\gamma}_i$ is the true 2D observation and $\Delta \gamma_i$ is the error introduced by noise, {\em i.e.}, 
\begin{align}
    \hat{\gamma}_j^T E_{ij} \hat{\gamma}_i = 0, \; (\gamma_j - \Delta\gamma_j)^T E_{ij} (\gamma_i - \Delta\gamma_i) = 0.
\end{align}
Minimizing reprojection errors $\Delta\gamma_i$ and $\Delta\gamma_j$ subject to this constraint solves the optimal estimate
\begin{align}
     (\Delta \gamma^*_i, \Delta \gamma^*_j) = \argmin_{(\gamma_j - \Delta\gamma_j)^T E_{ij} (\gamma_i - \Delta\gamma_i) = 0}(||\Delta \gamma_i||^2 + ||\Delta \gamma_j||^2).  \nonumber
\end{align}
Using Lagrange multipliers and notation $\Delta \gamma^T_i = (u_i, v_i,0)$ the problem becomes 
\begin{align}
     &(u_i^*,v_i^*,u_j^*,v_j^*,\lambda^*) = \argmin_{u_i,v_i,u_j,v_j,\lambda^*}  (u_i^2 + v_i^2 + u_j^2 + v_j^2)  \nonumber \\ \nonumber & + \lambda (\gamma^T_j - [u_j\;v_j\;0])
    E_{ij} (\gamma_i - 
    [u_i\;v_i\;0]^T). \nonumber 
\end{align}
This can be solved by differentiating with respect to the five variables and setting to zero. Specifically,
\begin{align}
    \left\{\begin{matrix}
    2u_i - \lambda (\gamma^T_j-[u_j\;v_j\;0])E_{ij}[1\;0\;0]^T=0 \\ 
    2v_i - \lambda (\gamma^T_j-[u_j\;v_j\;0])E_{ij}[0\;1\;0]^T=0 \\ 
    2u_j - \lambda [1\;0\;0]E_{ij}(\gamma^T_i-[u_j\;v_j\;0]^T)=0 \\ 
    2u_j - \lambda [1\;0\;0]E_{ij}(\gamma^T_i-[u_j\;v_j\;0]^T)=0 \\ 
    (\gamma^T_j-[u_j\;v_j\;0])E_{ij}(\gamma^T_i-[u_j\;v_j\;0]^T)=0
    \end{matrix}.\right.
\label{eq:constraints}
\end{align}
This is a set of five multi-linear polynomial equations in five unknowns. Setting the first derivative with respect to the five variables gives a $5 \times 5$ polynomial system. This system can be solved using HC without any further effort. Traditionally, however, the system is solved by eliminating four of five variables, gives a single 6-th order polynomial~\cite{hartley1997triangulation}. This gives excellent results but it is slow prompting~\cite{kanatani2008triangulation,lindstrom2010triangulation} to use an iterative method which is faster but is prone to being stuck in local minima. 

The N-view triangulation is not as well-explored despite the formulation of minimizing reprojection error is identical
\begin{align}
    & (\Delta \gamma^*_1, \Delta \gamma^*_2, \cdots, \Delta \gamma^*_N) = \\ \nonumber & \argmin_{\footnotesize \makecell{\Delta \gamma_1, \Delta \gamma_2,...,\Delta \gamma_N\;such\;that\\\forall i,j 
    (\gamma_j - \Delta\gamma_j)^T E_{ij} (\gamma_i - \Delta\gamma_i) = 0 }} \sum_{k=1}^{N} |\Delta \gamma_k|^2,
\end{align}
or
{\footnotesize
\begin{align}
     & (u^*_1,v^*_1,u^*_2,v^*_2,\cdots,u^*_N, v^*_N,\Lambda^*) =
     \\ \nonumber 
     &\argmin_{u^*_1,v^*_1,u^*_2,v^*_2,\cdots,u^*_N, v^*_N,\Lambda} \sum_{k=1}^{N} [(u^2_k + v^2_k)  + \\ \nonumber 
     & \sum^N_{i=1} \sum^N_{j=i+1} \lambda_k (\gamma^T_j - [u_j\;v_j\;0])
    E_{ij} (\gamma_i - \begin{bmatrix}
    u_i \nonumber \\
    v_i \nonumber \\
    0\nonumber 
    \end{bmatrix}),
\end{align}
}\\
\noindent where $\Lambda = \{\lambda_{ij} | i=1,2,...,N, j=i+1,...,N \}$. Note that there are $2N + \frac{N(N-1)}{2} = \frac{N^2 + 3N}{2}$ unknowns and setting first derivatives to zero gives $5 \times 5$, $9 \times 9$ and $14 \times 14$, for two, three, and four views, respectively, becoming exponentially more difficult to use with Gr\"obner basis and other traditional methods.~\cite{kukelova2013fast} restricts the consideration of all sequential pairwise essential matrices to these with the previous view, {\em i.e.}, $E_{12}$, $E_{23}$, {\em etc.} and uses the elimination template method with a $274 \times 305$ template;~\cite{larsson2017efficient} reduce the size of the  elimination template to $239 \times 290$. Note that the full problem gives an elimination template of $1866 \times 1975$ which is not practical to solve. Similarly, the 4-view triangulation gives improved error but it leads to large polynomial systems which are simply impractical to solve. Homotopy Continuation, however, can solve these problems and with improved efficiency, Table~\ref{tab:VisionProblems}.

\noindent \textbf{Trifocal relative pose estimation with unknown focal length} aims to estimate the relative pose between three views as well as the focal length. Trifocal pose estimation has drawn more attention recently~\cite{leonardos2015metric,chen2016trifocal, julia2017critical, kileel2017minimal, fabbri2020trplp}. These approaches often assume that the intrinsic camera calibration is available. Recently,~\cite{larsson2020calibration} estimates trifocal tensor with radial distortion, a minimal problem of one pinhole camera and two radial cameras. We now consider another minimal problem with three pinhole cameras with common unknown focal length. This minimal problem needs only 4 points correspondences across three views. Let the calibration matrix be $K = diag(f,f,1)$, where $f$ is the focal length. Consider three corresponding points ($\gamma_1$, $\gamma_2$, $\gamma_3$) in image ($1$,$2$,$3$), respectively, with unknown depth ($\rho_1$, $\rho_2$, $\rho_3$) respectively. Then, denoting ($R_{12}$, $T_{12}$) and ($R_{13}$, $T_{13}$) the relative pose of the second and third cameras with respect to the first, respectively, we have
\begin{align}
\left\{
\begin{matrix}
\rho_2 K^{-1} \gamma_2 = \rho_1 K^{-1} R_{12} \gamma_1 + \hat{T}_{12} \\
\rho_3 K^{-1} \gamma_3 = \rho_1 K^{-1} R_{13} \gamma_1 + T_{13},
\end{matrix}
\right.
\label{Eq:elimiationUnknownf}
\end{align}
where $\hat{T}_{12}$ is taken to have unit length. Thus, there are 11 poses and 3 depth unknowns. Since there are four sets of correspondences, there are in total one focal length, 11 pose and 12 depth unknowns, for a total of 24 unknowns. There are also four sets of vector Equation~\ref{Eq:elimiationUnknownf} which each gives 6 equations. If $R_{12}$ and $R_{13}$ are represented by quaternions there are a total of 26 equation in 26 unknowns which can be solve by HC. 

Alternatively, $\rho_2$ and $\rho_3$ can be written in terms of $\rho_1$ as 
\begin{align}
    \left\{
\begin{matrix}
\rho_2 e_3^TK^{-1} \gamma_2 = e_3^T\rho_1 K^{-1} R_{12} \gamma_1 + e_3^TK^{-1}T_{12} \\
\rho_3 e_3^T K^{-1} \gamma_3 = e_3^T\rho_1 K^{-1} R_{13} \gamma_1 + e_3^TK^{-1}T_{13}.
\end{matrix}
\right.
\end{align}
Substituting these back into Equation~\ref{Eq:elimiationUnknownf} gives 4 equations for each triplet of corresponding s for a total of 16 equations. The unknowns are 1 focal length, 11 pose and 4 depth unknowns for a total of 16 unknowns. If $R$ is represented as a quaternion one additional unknown and one additional equations is introduced per rotation matrix giving a total of 18 polynomial equations in 18 unknowns. This minimal problem cannot be solved using elimination template since the required memory is unavailable even on a high performance computing machine. However, our HC implementation can solve this system with 1376ms in CPU and 154.19ms in GPU, Table~\ref{tab:VisionProblems}.

\section{GPUs and Computer Vision}
\label{sec:GPU}

GPUs are often preferred over CPUs because of
their superior computational power, memory bandwidth, and energy
efficiency. For example, a V100 GPU provides an FP64 compute peak of 7 TFlop/s and memory bandwidth of
900 GB/s at 250 Watts. While one CPU core is faster and provides wider instruction sets, GPUs have many more cores, {\em e.g.}, 5,120 in the V100.
The key to unlocking the computational power of the GPU is to design algorithms that are highly parallel and use efficiently
all the cores.
%------------------------------------------------------------
\begin{figure}
    \centering
    \includegraphics[width = 0.7 \linewidth]{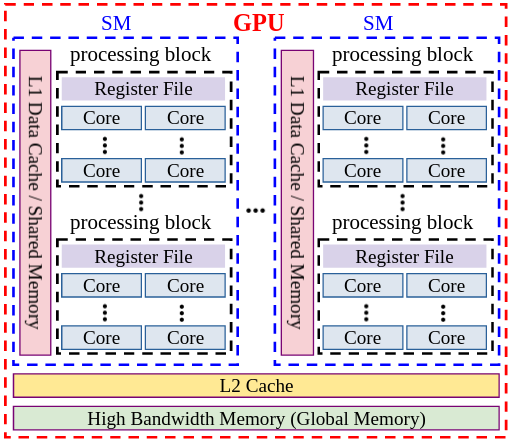}
    \caption{A schematic view of a modern NVIDIA GPU.}
    %\caption{The NVIDIA’s GPU architecture and memory hierarchy. In Tesla V100 GPU, there are 80 SMs  each  partitioned into four processing blocks, each having 16 cores. The register files per SM, shared memory/L1 cache per SM, L2 cache, and global memory are respectively 256KB and 96KB, 6,144KB and 16GB, with hit latencies of 3 $ns$, 22.68 $ns$, 156.33 $ns$, and roughly 366 $ns$, respectively~\cite{jia2018dissectVoltaGPU}.}
    \label{fig:GPUdiagram}
\end{figure}

Figure~\ref{fig:GPUdiagram} shows the GPU architecture. The CUDA cores are organized into Streaming Multiprocessors (SMs) where each SM has a number of CUDA cores. The GPU work is organized into \emph{kernels} that have two levels of nested parallelism - 
a coarse level that is data parallel and is spread across the SMs, 
and a fine level within each SM. 
The parallelism is organized in terms of thread blocks (TBs). A	TB
is scheduled for execution on one of the SMs and is data parallel with respect to the other TBs. Each TB is composed of multiple threads running in groups of $32$ called \emph{warps}. Threads in a TB can share data through a shared memory module. Private variables that have the scope of one thread are usually stored in the register file.
Algorithms must	be designed to support this type of parallelism.
%,
%and by doing it to use all the hardware. After a design for
%parallelism, next step is to maximize the bandwidth of reading
%inputs and writing results back to the global memory. 
%Another key to high performance is ensuring that reads/writes to/from the global memory are "coalescent". That is, threads in a warp always access a contiguous piece of memory in order to maximize the memory bandwidth~\cite{guide}.  
%That is, a warp access global memory at the same time, 
%e.g., by having consecutive
%threads access consecutive memory addresses that are also properly
%aligned~\cite{guide}.

%The memory hierarchy allows algorithms to reach the compute peak of	the GPU	if a certain amount of data reuse is achieved. For example, if the data is used only once, the 900 GB/s transfer rate limits the peak performance to 112.5 GFlop/s in double-precision (8 bytes per element). Thus, to reach 7 TFlop/s, an algorithm would need about 64 times data reuse, {\em i.e.}, each 8 bytes loaded into fast memory to be used in 64 FLOPs. This can be achieved for General Matrix-Matrix Multiplications (GEMM) for example~\cite{ntd10}, and many dense linear algebra algorithms such as the ones in LAPACK\footnote{http://www.netlib.org/lapack}, and subsequently MAGMA (LAPACK for GPUs), that can be expressed using GEMMs and BLAS\footnote{http://www.netlib.org/blas} in general~\cite{tdb10}. 
The multi-level memory hierarchy enables compute-bound operations, like general matrix-matrix multiplication (GEMM), perform close to the compute peak of the GPU~\cite{ntd10}. Subsequently, many dense linear algebra algorithms such as the ones in LAPACK,
%\footnote{http://www.netlib.org/lapack}, 
and subsequently MAGMA (LAPACK for GPUs), that can be expressed using GEMMs and BLAS
%\footnote{http://www.netlib.org/blas} 
in general~\cite{tdb10}. Some numerical algorithms, like the one mentioned in Section~\ref{sec:hc_gpu}, involve many independent computations ({\em e.g.} dense factorizations) on relatively small matrices. These algorithms, are limited by the memory bandwidth, but have a high degree of parallelism, which is suitable for GPUs. Maximizing data reuse is possible by caching each matrix entirely in the register file or shared memory, which enables GPUs to outperform multicore CPUs in these types of algorithms. \\
\indent Algorithms in computer vision are naturally data-parallel and
computationally	intensive, and therefore a good fit for modern GPUs. Computational patterns involving one-to-one mappings like an image being	modified by different
filters, can benefit from the data parallelism and the memory
hierarchy to chain the applications of filters. Many-to-one mappings
that involve summations of certain buffers also can
benefit the memory hierarchy and do	it fast	in MPs,	vs. creating 
a sequence of fragment programs to simulate summations in older 
GPUs without memory hierarchy. CNNs for example map many 
convolutions (involving summation) on many images into batched
GEMMs~\cite{cudnn}. Many-to-many computational patterns 
can be mapped efficiently to GPUs as well~\cite{pharr05}.
Operations that	can not	be mapped efficiently to GPU have been left
in general for the CPUs. This usually involves irregular computations
on small data sets where there is not enough parallelism, and
computations with a lot	of data	dependencies (like	solving	a small
system of equations). Still, techniques like batching
computations to increase parallelism and developments in numerical
linear algebra libraries for GPUs~\cite{ahmad_isc16,batched-onesided},
have laid the groundwork for many 
more algorithms to be easily ported and benefit GPU use. 
Often algorithms that
have been avoided before due to their computational cost are
becoming preferred for GPUs when current advances make their GPU
mapping very efficient.	This is the case for the HC that we target to develop and illustrate in this paper.

\section{GPU Implementation of HC}
\label{sec:hc_gpu}
\indent The homotopy continuation process can be parallelized in two ways: First, observe that since HC follows many independent tracks to convergence, a straightforward approach would be to assign each track to a thread. However, the efficiency of GPU processing depends on \emph{(i)} number of threads processing many tracks in parallel, and \emph{(ii)} avoiding costly data transfer rates by using the fast register files, or at least L1 caches v.s. the slower L2 cache or even slower global memory, Figure~\ref{fig:GPUdiagram}. In our application, each track requires a few Kbytes while the available memory is 125, 46, 37, and 97K bytes for register file, L1 cache, L2 cache, and global memory, respectively, for one thread per track. Thus any process requiring more than 125 + 46 + 37 = 208 bytes of memory is forward to use the very slow global memory. As a result, each track must make use of many threads, and not only the processing must be parallelized, but so must the use of memory with the aim of keeping everything in register file, shared memory, or at least L1 cache. \\
\indent Observe that the other extreme of spreading a trade over numerously many threads starts becoming counterproductive because the synchronization of threads employs the slower shared memory (2 clock cycles per 32 threads) so that if 2048 threads are employed per track, 128 clock cycles ($\sim$104 ns) times the tens of thousands of their synchronization is needed which becomes an unnecessary overhead. \\
\indent The optimal balance for the target applications is to assign a track to a warp (32 threads) using one GPU core. This gives the application access to $256K / 64 = 4K$ very fast register file memory and $96K / 64 = 1.5K$ of fast L1 cache (if no shared memory is used), well satisfying the memory requirement of the target application. On the other hand, the cost of thread synchronization is only 2 clock cycles.

\indent The second intuition aims for parallelizing HC within each warp by \emph{(i)} solving a system of linear equations in both the prediction step, Eq.~\ref{Eq:linsysRK} and the correction step, Eq.~\ref{Eq:homotopy}, and \emph{(ii)} evaluating the Jacobian matrix $\partial H/ \partial x$, $\partial H/ \partial t$, and the homotopy $H$, Eq.~\ref{Eq:prediction} and~\ref{Eq:newton}.

\noindent \textbf{Linear System Solver:} The vast majority of work on solving linear systems on GPU is centered around large matrices, motivating a hybrid CPU+GPU approach~\cite{hybridSysSolver2010, denselinsys2010_magma}. For smaller matrices like ours, cuBLAS or MAGMA~\cite{ahmad2018magma,plasma-magma,magma} can be used. A linear system is generally solved by an LU factorization with partial pivoting followed by two triangular solves. 
%The LU factorization of a matrix typically proceeds by summing the first row and others to zero out all but the first row in the first column, then proceed to zero out all but the first two rows in the second column, etc. 
%The LU factorization of a matrix proceeds one column at a time. For each column, \emph{(i)} a \emph{pivot} is chosen based on the maximum absolute value, \emph{(ii)} a row interchange is applied so that the pivot element is brought on the diagonal, \emph{(iii)} the current column is scaled with respect to the pivot, and \emph{(iv)} a rank-1 update is applied to the trailing matrix. 
%Note that assigning a thread per element is quite wasteful as some threads have to wait a long time before others are done. 
The LU factorization in MAGMA is fast, typically $15\%$ to $80\%$ faster than cuBLAS for small matrices.
%namely 12$\mu s$ and 38$\mu s$ for $4\times4$ and $20\times20$ matrices, much faster than cuBLAS. 
%This is because for matrices less than $32\times32$, MAGMA assigns one row per thread in a warp, and the factorization proceeds one column at a time in the register file, while inter-thread communication occurs in shared memory. 
However, we found out that cuBLAS is faster than MAGMA for the combined (factorization + solve) operation. This is mainly due to a slow triangular solver kernel in MAGMA, which does not take advantage of small matrices. \\
%In terms of overall times to solve both steps, cuBLAS is faster than MAGMA. 
\indent Our contribution to improving these standard libraries for our purposes is twofold. First, solving the linear system as two separate GPU kernels causes redundant global memory traffic. The two kernels can be fused into one if the matrices are small, thus maximizing data reuse in the register file. The proposed \emph{kernel fusion} significantly speeds up the solution. Second, in solving a linear system $Ax=b$, the decomposition can act on the augmented matrix $[A\;b]$, which implicitly carries out the triangular solve with respect to the $L$ factor of $A$. 
%so that as $A$ is being transferred to the triangular matrix $U$, so is $b$, effectively factoring out the $L$ portion which is now done in the decomposition process. 
The second triangular solve uses the cached $U$ factor after the factorization is complete. The proposed fused kernel is now integrated into the MAGMA library.

\noindent \textbf{Parallel Evaluations of the Jacobian and Vectors:}

The main bottleneck to parallel evaluations of the elements of the Jacobian matrix $\partial H/ \partial x$ and the vectors $\partial H/ \partial t$ and $H$ is the heterogenuity of its elements which prevents evaluation by many threads requiring a uniform format. This heterogenuity can be illustrated by a simple example of a system with two variables $X = (x_1, x_2)$ where the Jacobian elements are spanned by monomials, for example, A = $2a_1x_1+4a_2x_1x_2+8a_3x_2^2$ or B = $5a_4x_1x_2+7a_5x_2^2$, where the coefficients $a_i$ are linear interpolation of corresponding elements in the start and target systems. The straightforward approach to homogenize these expressions is to write each as a sum over all possible monomials and associate a scalar zero with those absent from the Jacobian elements in parallel. However, due to the extreme sparsity, the process is highly inefficient.\\
\indent Alternatively, consider $K$ the maximum number of terms in the Jacobian matrix elements; in the above examples, A has three terms and B has two terms, so that $K=3$ if these were the only elements of the Jacobian matrix. Furthermore, consider that each term consists of a scalar multiplied with a coefficient and a number of variables, \emph{e.g.}, the third term of A is a product of $(8,a_3,x_3,x_3)$ while the first term of B is $(5,a_4,x_1,x_2)$. Note that the first term of A is a product of $(2,a_1,x_1)$. Thus, to homogenize the expression, it is written as $(2,a_1,x_1,x_3)$ where the auxiliary variable $x_3=1$. Now all terms of both A and B can be written as
\begin{align}
    U = \sum\nolimits_{k=1}^{K}s_k a_{k,j} x_{k,m_1} x_{k,m_2} \cdots x_{k,m_M}, \nonumber
\end{align}
where $s_k$ is a scalar, $a_{k,j}$ identifies a coefficient, $x_{k,m_i}$ identifies one of the variables, including $x_3=1$, and $M$ is the maximal number of variables in a term. With this in mind the only data to be communicated for the parallel computation of $U$ is $(s_k,a_{k,j}, x_{k,m_1},x_{k,m_2},...,x_{k,m_M})$ where $a_{k,j}$, $x_{k,m_i}$ are pointers to data stored in shared memory and accessed by an index, \emph{i.e.}, A is represented by $((2,1,1,3),(4,2,1,2),(8,3,2,2))$ and B is represented by $((5,4,1,2),(7,5,2,2),(0,1,1,1))$. Note that $\partial H/ \partial t$ and $H$ are evaluated in the same way although the coefficients $a_k$ are different. This homogeneous form allows for parallel computation of all elements of the Jacobian matrix $\partial H/ \partial x$ and the vectors $\partial H/ \partial t$ and $H$. \\
\indent Finally, there is an issue on how to allocate the parallel computations per thread. Recall that each track is assigned to a warp which has 32 threads. Since the matrices are generally less than $32\times32$, and since the subsequent operation of LU decomposition is row-by-row with one thread per row, it makes sense to assign one row per thread.
\begin{figure}[!ht]
    \centering
    \includegraphics[width=0.8\linewidth]{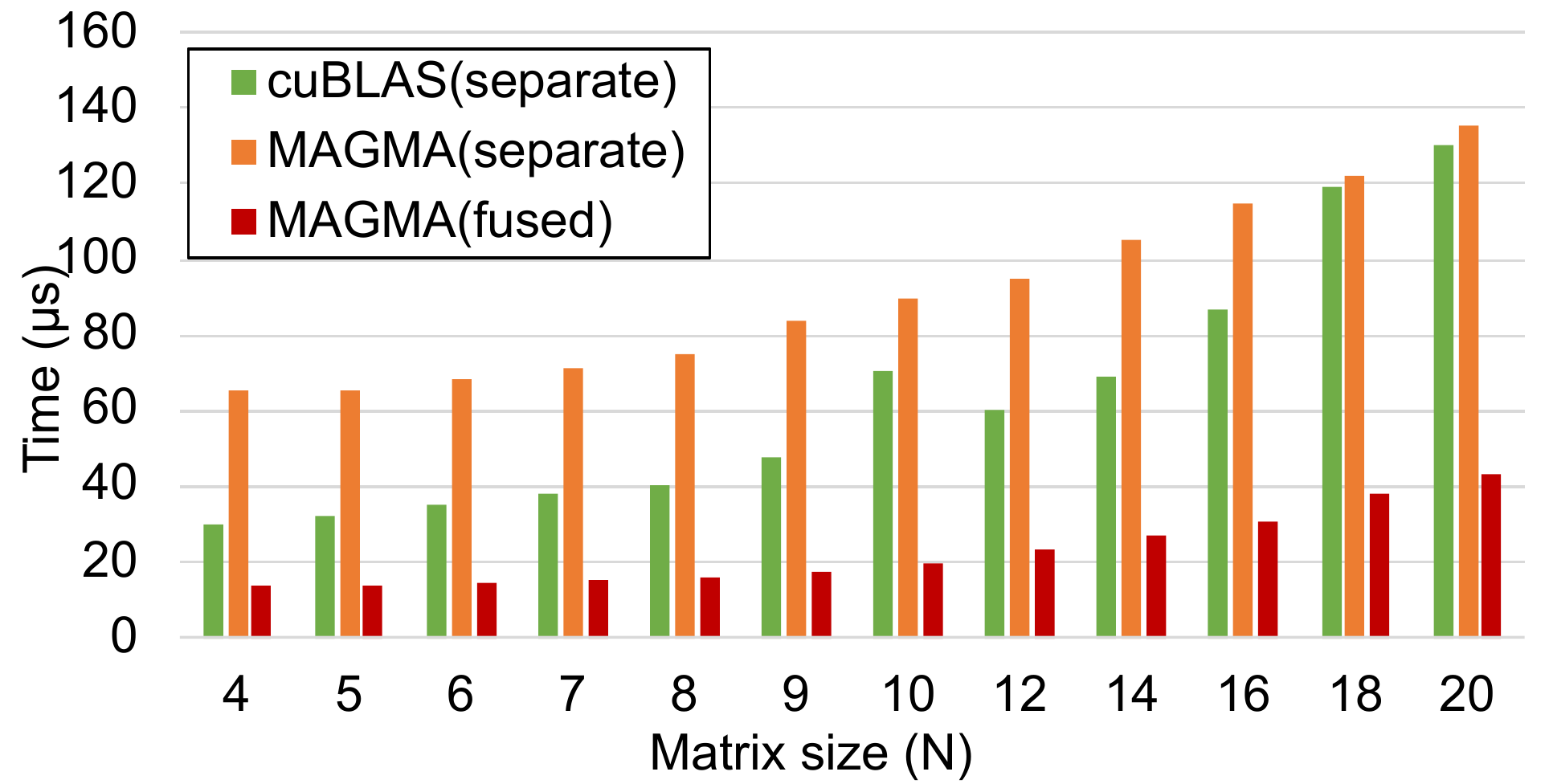}
\caption{The performances of the batch linear systems of MAGMA with kernel fusion,  MAGMA, and cuBLAS.}
\label{fig:cgesv_v100}
\end{figure}
\begin{figure}[!ht]
    \centering
       (a) \includegraphics[width=0.42\linewidth]{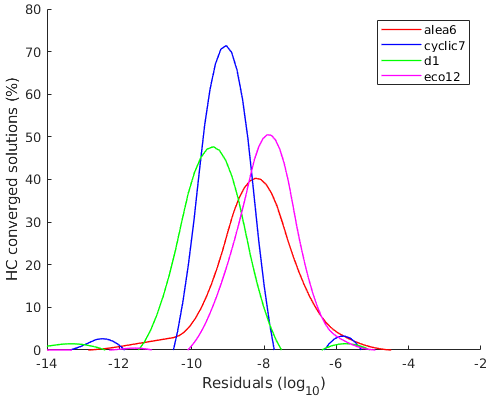}
    (b)\includegraphics[width=0.42\linewidth]{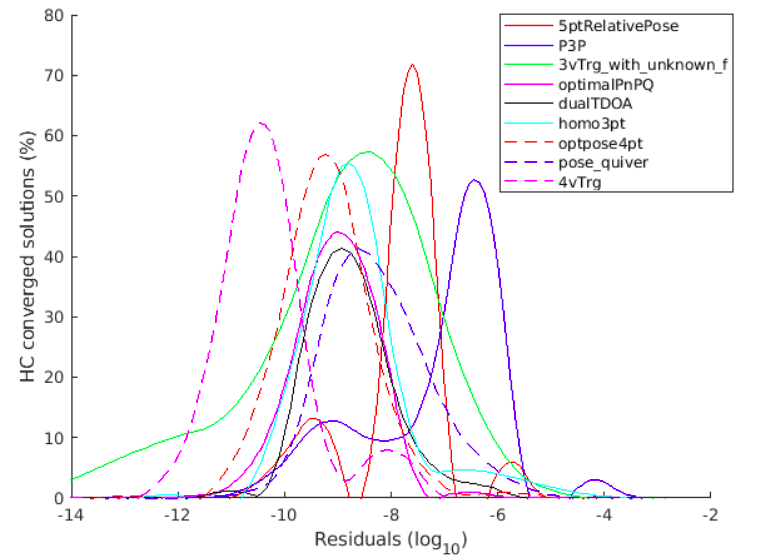}
\caption{Numerical accuracy of GPU-HC on (a) polynomial system benchmarks (b) computer vision problems.}
\label{fig:BenchMarkingProblemCurve}
\end{figure}

\section{Experiments}
\label{sec:experiments}
The experiments aim at testing kernel fusion for batch linear systems, and measuring performances on polynomial system benchmarks as well as computer vision problems. We use an 8-core 2.6GHz Intel Xeon CPU and an nVidia Quardro RTX 6000 GPU, unless otherwise specified.\\

\noindent \textbf{Kernel-Fused Batch Linear Systems:}
The performance of the batched linear systems with kernel fusion and augmented matrix, Section~\ref{sec:hc_gpu}, is compared with cuBLAS and MAGMA in Figure 3 on a Tesla V100-PCIe GPU for 1000 matrices with sizes ranging from $4\times4$ to $20\times20$. Evidently, kernel-fused MAGMA outperforms cuBLAS with speedup of $2.23\times$ to $3.65\times$ and MAGMA with speedups ranging from $3.11\times$ to $4.91\times$. \\

\noindent \textbf{Polynomial System Benchmarks:}
We selected four representative benchmark systems~\cite{parisse2013probabilistic, backelin1991we, hong1994safe, morgan2009solving} to evaluate the performance of our GPU-HC. Table~\ref{tab:benchmark} shows GPU-HC speedup ranging from $10\times$ to $26\times$. Figure~\ref{fig:BenchMarkingProblemCurve} (a) shows that the speedup is not at the cost of lower accuracy, \emph{i.e.}, the GPU-HC computed solutions satisfy the polynomial system with high accuracy. 
%The full polynomial formulations of the benchmark systems are included as a supplement. 
\\
\begin{table}[!ht]
    \centering
    \begin{tabular}{ccccccc}
        \hline
        {\footnotesize\textbf{Problems}} & {\footnotesize\makecell{\textbf{\# of} \\  \textbf{Unkns.}}} & {\footnotesize\makecell{\textbf{\# of} \\ \textbf{Sols.}}} & {\footnotesize\makecell{\textbf{CPU} \\ \textbf{(ms)}}} & {\footnotesize\makecell{\textbf{GPU}\\  \textbf{ (ms)}}} & $\frac{\textbf{CPU}}{\textbf{GPU}}$\\
        \hline
        {\footnotesize Alea-6~\cite{parisse2013probabilistic}} & \footnotesize 6 & \footnotesize 387 & \footnotesize 156.71 & \footnotesize 5.94 &  \footnotesize 26.39$\times$ \\
        {\footnotesize Cyclic-7~\cite{backelin1991we}} & \footnotesize 7 & \footnotesize 924 & \footnotesize 219.02 & \footnotesize 8.35 & \footnotesize 26.23$\times$  \\
        {\footnotesize D-1~\cite{hong1994safe}} & \footnotesize 12 & \footnotesize 192 & \footnotesize 100.57 & \footnotesize 7.09 & \footnotesize 14.18$\times$ \\
        {\footnotesize Eco-12~\cite{morgan2009solving}} & \footnotesize 12 & \footnotesize 1024 & \footnotesize 279.12 & \footnotesize 27.43 & \footnotesize 10.18$\times$ \\
        \hline
    \end{tabular}
    \caption{Performance of GPU-HC on benchmark problems.}
    \label{tab:benchmark}
\end{table}

\noindent \textbf{Computer Vision Problems:}
We consider a sample of minimal problem in computer vision ranging from the classic pose estimation P3P to the more complex 3-view triangulation as well as two problems that have not been explored previously: {\em (i)} 4-view triangulation is an extension of 3-view triangulation~\cite{byrod2007fast}. As far as we know, this is the first attempt to explore this problem. {\em (ii)} trifocal relative pose estimation without focal length is an extension of existing trifocal relative pose estimation problems~\cite{fabbritrifocal, larsson2020calibration} to the uncalibrated scenario; as far as we know, this problem has not been explored previously.
The most popular technique for solving polynomial system is the elimination template approach~\cite{larsson2017efficient} and is used to gauge the performances of GPU-HC. Table~\ref{tab:VisionProblems} shows a comparison of the elimination template performances with that of HC on CPU and on GPU. Each problem is instantiated 20 times with random parameters and its performance is averaged. The start systems for HC are generated with monodromy module in Macaulay2~\cite{duff2021applications}. Numerous factors affect the speedup  of GPU-HC over CPU-HC, including the number of solutions, number of unknowns, and the number of terms in the polynomial evaluations of the Jacobian matrix. The performances of elimination template is dependent on the size of the linear system it solves which itself is related to the number of solutions of the polynomial system. Note that for the top three problems the elimination template cannot compute the basis of the quotient ring of the system even with ample memory. A review of Table~\ref{tab:VisionProblems} which is ordered by elimination template time, shows that with the exception simpler problem such as P3P (the bottom four rows), the GPU-HC outperforms the elimination template. GPU-HC opens the door to more complex problems that the elimination template cannot handle. %Additional experimental data can be found in the supplementary materials.
\\

\begin{table*}[htbp]
\small
\centering
\begin{tabular}{cccccccc}
\hline
\footnotesize \textbf{Problems} & \footnotesize \textbf{\makecell{$\#$ of \\ Unkns.}} & \footnotesize \textbf{\makecell{$\#$ of \\Sols.}} & \footnotesize \textbf{\makecell{Elim. Temp. \\ (ms)}} & \footnotesize \makecell{\textbf{CPU}\\\textbf{(ms)}} & \footnotesize \makecell{\textbf{GPU}\\\textbf{(ms)}}& \large $\frac{\textbf{CPU}}{\textbf{GPU}}$ & \large $\frac{\textbf{Elim. Temp.}}{\textbf{GPU}}$ \\
\hline
\makecell{trifocal rel. pose, unknown focal length} & 18 & 1784 & X & 1456.22 & \textbf{154.19} & 9.44$\times$ &  N./A. \\
\hline
\makecell{4-view triangulation} & 14 & 296 & X & 156.28 & \textbf{18.60} & 8.32$\times$ & N./A. \\
\hline
\makecell{5 pt rel. pose \& depth recon.} & 16 & 160 & X & 150.94 & \textbf{26.89} & 5.61$\times$  & N./A.  \\
\hline
\makecell{6 pt rolling shutter abs. pose w. 1-lin.~\cite{albl2019pami}} & 18 & 160 & X & 158.48 & \textbf{27.11} & 5.85$\times$ & N./A. \\
\hline
\makecell{3-view triangulation~\cite{byrod2007fast}} & 9 & 94 & 612.432 & 101.86 & \textbf{8.17} & 24.19$\times$ & 38.24 $\times$ \\
\hline 
\makecell{optimal PnP with quaternion~\cite{nakano2015globally} } & 4 & 128 & 36.329 & 80.26 & \textbf{7.18} & 11.18$\times$ & 5.06$\times$ \\
\hline
P4P, unknown focal length \& radial distortion~\cite{bujnak2010new} & 5 & 192 & 9.03 & 130.79 & \textbf{7.51} & 17.42$\times$ & 1.2$\times$ \\
\hline
\makecell{2-view triangulation with radial distortion~\cite{kukelova2019cvpr}} & 5 & 28 & 5.92 & 66.22 & \textbf{3.06} & 21.64$\times$ & 1.93$\times$ \\
\hline
optimal P4P abs. pose~\cite{svarm2016city} & 5 & 32 & 1.864 & 53.13 & \textbf{1.57} & 33.84$\times$ & 1.19$\times$ \\
\hline
3 pt rel. pose w. homography constraint~\cite{saurer2014homography} & 8 & 8 & 1.472 & 51.25 & \textbf{0.95} & 53.95$\times$ & 1.55$\times$ \\
\hline
\makecell{PnP, unknown principal point~\cite{larsson2018camera}} & 10 & 12 & \textbf{1.466} & 58.31 & 3.87 & 15.07$\times$ &  0.38$\times$ \\
\hline
rel. pose w. quiver, unknown focal length~\cite{kuang2013pose} & 4 & 28 & \textbf{1.082} & 56.01 & 1.23 & 45.54$\times$ & 0.88$\times$ \\
\hline
P3P abs. pose~\cite{kukelova2008automatic} & 3 & 8 & \textbf{0.063} & 39.64 & 0.22 & 180.18$\times$ & 0.29$\times$  \\
\hline
\makecell{5 pt rel. pose w.o. depth recon.~\cite{nister2004efficient}} & 3 & 27 & \textbf{0.035} & 55.48 & 0.96 & 57.79$\times$ & 0.036$\times$  \\
\hline
\multicolumn{8}{l}{X: it is impossible for elimination template to solve because of an out of memory issue} \\
\hline
\end{tabular}
%\captionsetup[table]{skip=5pt}
\caption{Performance of GPU-HC and CPU-HC {\it vs.} elimination template in application to several minimal problems.}
\label{tab:VisionProblems}
\end{table*}

\noindent{\bf{Conclusion:}} We presented GPU-HC, a GPU implementation of HC that is generic and can be easily applied to any computer vision problem formulated as a system of polynomial equations. The significant speedup of GPU-HC is an enabler in that HC can now be efficiently used for moderately complex problems in place of completing approaches. GPU-HC also enables the exploration of problems whose complexity has thus far evaded a practical solution. 

{\small
\bibliographystyle{ieee}
\bibliography{egbib,magma_ref}
}

\clearpage

% =================== SUPPLEMENTARY MATERIALS =====================
\title{ \textbf{GPU-Based Homotopy Continuation for Minimal Problems in Computer Vision: Supplementary Material}}
\author{Chiang-Heng Chien\\
School of Engineering\\
Brown University\\
{\tt\small chiang-heng\_chien@brown.edu}
\and
Hongyi Fan\\
School of Engineering\\
Brown University\\
{\tt\small hongyi\_fan@brown.edu}
\and
Ahmad Abdelfattah\\
Innovative Computing Laboratory\\
University of Tennessee\\
{\tt\small ahmad@icl.utk.edu}
\and
Elias Tsigaridas\\
INRIA\\
{\tt\small elias.tsigaridas@inria.fr }
\and
Stanimire Tomov\\
Innovative Computing Laboratory\\
University of Tennessee\\
{\tt\small tomov@icl.utk.edu}
\and
Benjamin Kimia\\
School of Engineering\\
Brown University\\
{\tt\small benjamin\_kimia@brown.edu}
}
\maketitle

In this supplementary material, we aim to show the effectiveness of the proposed 4-view triangulation problem using a synthetic multiview dataset and a real dataset. Furthermore, we also release full polynomial formulations of all the tested problems in: \url{https://anonymous.4open.science/r/Minimal-Problems-in-Computer-Vision-201D}.

\setcounter{section}{0}
% \section{Performance of GPU-HC, CPU-HC, and elimination template on additional minimal problems}

% To show the superiorities of the proposed GPU-HC, we include another two recently-published minimal problems: \emph{(i)} 2-view triangulation with radial distortion~\cite{kukelova2019cvpr}, and \emph{(ii)} rolling shutter absolute pose estimation using 6 points (R6P1lin)~\cite{albl2019pami}. The authors of both papers pointed out that although their proposed minimal problem can be solved by Gr\"obner basis method, it is too slow to be useful in practice. Therefore, they put a lot of efforts on reducing the minimal problems into smaller degree polynomials. On the other hand, using the proposed GPU-HC, the two minimal problems can be solved directly without any further efforts. Extended from Table 2 of the manuscript, the blue rows shown in Table~\ref{tab:VisionProblems} in this supplementary material are the performances of GPU-HC, CPU-HC, and elimination template of the two problems. In R6P1lin problem, the elimination template method cannot compute the basis of the quotient ring of the system even with ample memory, while in the 2-view triangulation with distortion, GPU-HC evidently outperforms CPU-HC and the Gr\"obner basis method. The additional experiments further demonstrate that the proposed GPU-HC not only opens the door to more complex problems that the elimination template cannot deal with, for some small problems GPU-HC can also perform faster than the elimination template method.

\section{Effectiveness of the Four-View \\ Triangulation Problem}
\begin{figure}[!ht]
    \centering
    \includegraphics[width = 0.9 \linewidth]{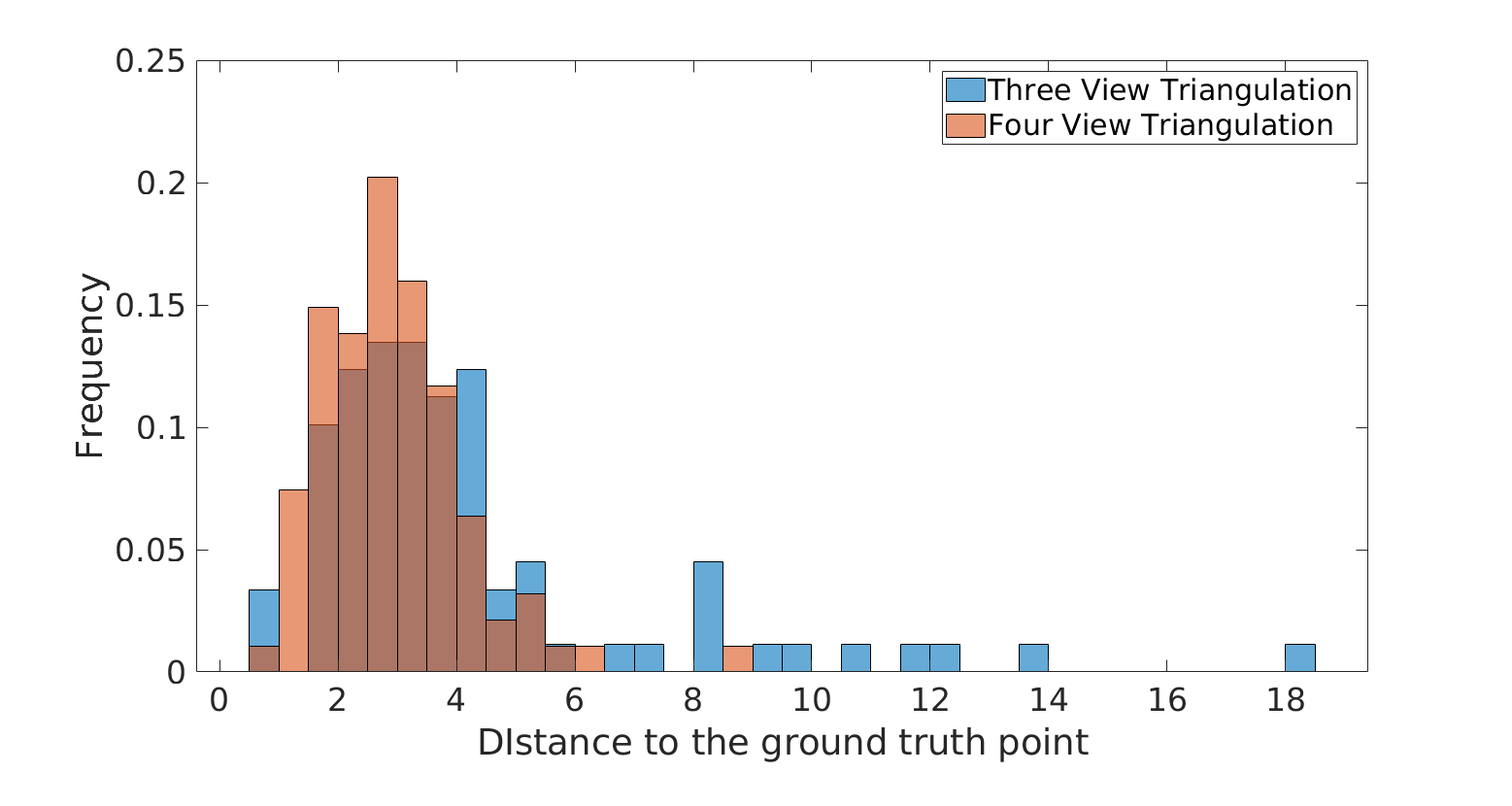}
    \caption{The 2D error histogram of 3-view and 4-view triangulation with synthetic Multiview dataset \protect\cite{fabbri2019trifocal}.}
    \label{fig:triSyn}
\end{figure}

To show the usefulness of 4-view triangulation method, synthetic multiview dataset from~\cite{fabbri2020trplp} is used for evaluation. We randomly select 200 points from random 4 pairs of views. To simulate the noise on the observed points and the calibration, the poses are perturbed with  $\mathcal{N}(0,2.0)$ and the observed points are perturbed with $\mathcal{N}(0,1.0)$. We compare our 4-view triangulate result with 3-view triangulation~\cite{kukelova2013fast}, Figure~\ref{fig:triSyn}. It is clear to see that the error between the optimized 2D points and the ground truths of 4-view triangulation is smaller than 3-view triangulation. Quantitatively, the mean error of 4-view triangulation is 2.9712 pixels, while in 3-view triangulation, the error is 4.2074 pixels. Therefore, adding one more view to do triangulation would drop the projection error significantly.

\begin{figure}[htp]
    \centering
    (a) \includegraphics[width=0.55 \linewidth]{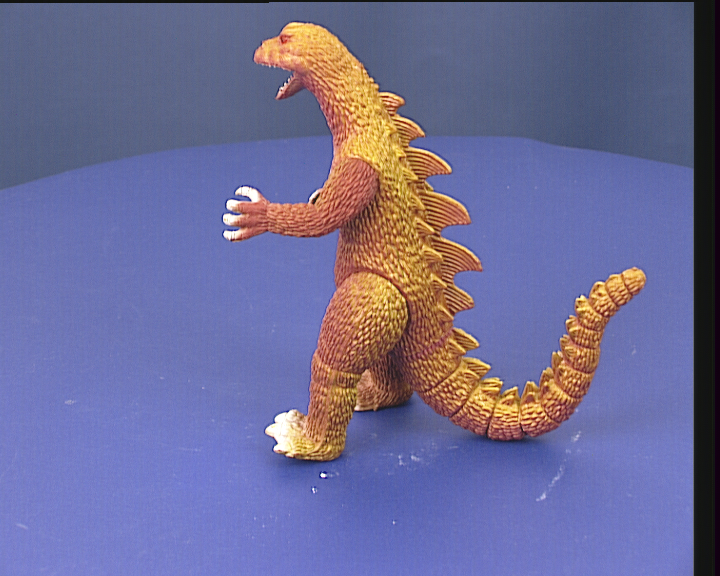}\\
    (b) \includegraphics[width=0.55 \linewidth]{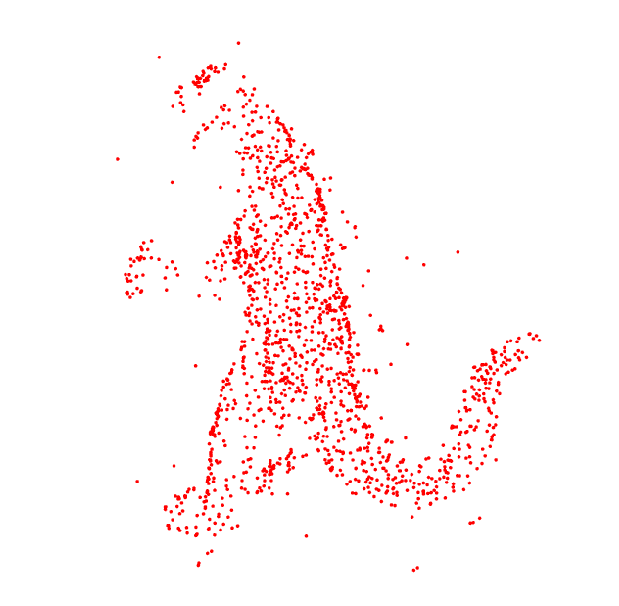}
    \caption{(a) Sample image of dinosaur sequence from VGG Multiview Dataset \protect\cite{dinasour}. (b) Triangulation result using GPU-HC 4-view triangulation method.}
    \label{fig:realTriangulation}
\end{figure}

Apart from using the synthetic dataset, a real dataset~\cite{dinasour} is also employed to challenge the effectiveness of the 4-view triangulation. We use the dinosaur sequence from~\cite{dinasour} which contains 36 images with 4,983 corresponding points. To deploy our 4-view triangulation method, only 1,516 points that are co-visible by more than 3 views are used. The triangulation result is shown in Figure~\ref{fig:realTriangulation}. Note that after GPU-HC computes the optimal positions of the image points, the projection lines from all four images always intersect in the 3D space. In such a case, we are free to select any two views to find the position of the 3D points given the camera extrinsic matrix of these two views. The processing time of this whole sequence using the proposed GPU-HC is around 10.6 seconds. 

\newpage

\end{document}